\documentclass[journal]{IEEEtran}

\IEEEoverridecommandlockouts                              %

\usepackage{tcolorbox}
\tcbuselibrary{theorems}
\newtcbtheorem{boxtheorem}{\it Theorem}{colback=gray!30, colframe=black!60, title=\textit}{thm}

\title{\LARGE \bf
PIETRA: Physics-Informed Evidential Learning for\\
Traversing Out-of-Distribution Terrain
}

\author{Xiaoyi Cai$^1$, James Queeney$^{1, 2}$, Tong Xu$^3$, Aniket Datar$^3$, Chenhui Pan$^3$,\\
Max Miller$^1$, Ashton Flather$^1$, Philip R. Osteen$^4$, Nicholas Roy$^1$, Xuesu Xiao$^3$, and Jonathan P. How$^1$%
\thanks{This research was partially sponsored by ARL grant W911NF-21-2-0150 and by ONR grant N00014-18-1-2832. JQ was exclusively supported by MERL. Supplementary Video: \url{https://youtu.be/OTnNZ96oJRk}}
\thanks{$^1$Massachusetts Institute of Technology, Cambridge, MA 02139, USA. {\tt\{xyc,\;queeney,\;maxm25,\;ashton\_f,\;nickroy,\;jhow\}@mit.edu}.}%
\thanks{$^2$Mitsubishi Electric Research Laboratories (MERL), Cambridge, MA 02139, USA. {\tt queeney@merl.com}.}%
\thanks{$^3$George Mason University, Fairfax, VA 22030, USA. {\tt\{txu25, adatar, cpan7, xiao\}@gmu.edu}.}%
\thanks{$^4$DEVCOM Army Research Laboratory, Adelphi, MD 20783, USA. {\tt\small philip.r.osteen.civ@army.mil}. Distribution Statement A: Approved for public release; distribution is unlimited.}%
}

\usepackage{amsmath,amsthm,amssymb, amsfonts} %
\usepackage{bbold}
\usepackage{mathtools}
\usepackage{graphicx}
\usepackage{xcolor}
\usepackage{pdfpages}
\usepackage{bm}
\usepackage{url}
\usepackage{enumerate}
\usepackage{float,balance}
\usepackage[sort, compress]{cite}
\usepackage{cases,balance}
\usepackage[pdftex, pdfstartview={FitV}, pdfpagelayout={TwoColumnLeft},bookmarksopen=true,plainpages = false, colorlinks=true, linkcolor=black, citecolor = black, urlcolor = blue,filecolor=black , pagebackref=false,hypertexnames=false, plainpages=false, pdfpagelabels ]{hyperref}

\usepackage[capitalise]{cleveref}

\usepackage[font=footnotesize]{caption}

\usepackage{array}
\usepackage{booktabs}
\usepackage{colortbl}
\usepackage{multirow} %
\usepackage{threeparttable} %

\usepackage{placeins}

\usepackage{comment} %
\excludecomment{comments} %

\usepackage{algorithm}
\usepackage[noend]{algpseudocode} %
\definecolor{commentclr}{RGB}{110, 149, 204}
\definecolor{deeppink}{rgb}{1.0, 0.08, 0.58}

\makeatletter
\newcommand\fs@spaceruled{\def\@fs@cfont{\bfseries}\let\@fs@capt\floatc@ruled
  \def\@fs@pre{\vspace{0.6\baselineskip}\hrule height.8pt depth0pt \kern2pt}%
  \def\@fs@post{\kern2pt\hrule\relax}%
  \def\@fs@mid{\kern2pt\hrule\kern2pt}%
  \let\@fs@iftopcapt\iftrue}
\makeatother

\usepackage{tikz}

\usepackage{tcolorbox}
\newtcbox{\dashedbox}[1][]{
  math upper,
  baseline=0.4\baselineskip,
  nobeforeafter,
  colback=white,
  boxrule=0pt,
  enhanced jigsaw,
  boxsep=0pt,
  top=2pt,
  bottom=2pt,
  left=2pt,
  right=2pt,
  borderline horizontal={0.5pt}{0pt}{dashed},
  borderline vertical={0.5pt}{0pt}{dashed},
  #1
}

\newcommand{\Param}{\boldsymbol{\psi}}
\newcommand{\setParam}{\boldsymbol{\Psi}}

\newcommand{\setX}{\mathbf{X}} %

\newcommand{\lcvar}[2]{{\text{CVaR}_{#1}^{\leftarrow}(#2)}}
\newcommand{\rcvar}[2]{{\text{CVaR}_{#1}^{\rightarrow}(#2)}}

\newcommand{\emdsq}{{\text{EMD}^2}}
\newcommand{\uemdsq}{{\text{UEMD}^2}}

\newcommand{\dir}{{\text{Dir}}}
\newcommand{\cat}{{\text{Cat}}}

\newcommand{\cumsum}{{\text{cs}}}

\newcommand{\E}{{\mathbb{E}}}

\newcommand{\R}{\mathbb{R}}

\usepackage[export]{adjustbox} %

\newcommand{\tr}{^\top}
\newcommand*{\defeq}{:=}

\theoremstyle{plain}%

\theoremstyle{remark}

\theoremstyle{definition}

\graphicspath{{Figs/}}

\begin{document}
\maketitle

\begin{abstract}
Self-supervised learning is a powerful approach for developing traversability models for off-road navigation, but these models often struggle with inputs unseen during training. Existing methods utilize techniques like evidential deep learning to quantify model uncertainty, helping to identify and avoid out-of-distribution terrain. However, always avoiding out-of-distribution terrain can be overly conservative, e.g., when novel terrain can be effectively analyzed using a physics-based model. To overcome this challenge, we introduce Physics-Informed Evidential Traversability (PIETRA), a self-supervised learning framework that integrates physics priors directly into the mathematical formulation of evidential neural networks and introduces physics knowledge implicitly through an uncertainty-aware, physics-informed training loss. Our evidential network seamlessly transitions between learned and physics-based predictions for out-of-distribution inputs. Additionally, the physics-informed loss regularizes the learned model, ensuring better alignment with the physics model. Extensive simulations and hardware experiments demonstrate that PIETRA improves both learning accuracy and navigation performance in environments with significant distribution shifts.
\end{abstract}

\section{Introduction}

Recent advancements in perception and mobility have accelerated the deployment of autonomous robots in challenging real-world environments such as office spaces, construction sites, forests, deserts and Mars~\cite{ginting2024seek, triest2024unrealnet, frey2024roadrunner, chung2024pixel, meng2023terrainnet, takemura2024uncertainty}, where both geometric and semantic comprehension of the terrain is crucial for reliable navigation. In these settings, self-supervised traversability learning has emerged as a powerful tool to train neural networks (NNs) to predict terrain models from navigation data without manual labeling~\cite{gasparino2024wayfaster, Frey-RSS-23, seo2023learning, cai2024evora}, where the learned representations can work directly with model-based motion planners, providing better interpretability and flexibility compared to fully learned navigation policies. However, the lack of abundant and diverse training data limits the reliability of learned traversability models in novel environments. This is a well-known issue in the learning literature that arises from out-of-distribution (OOD) inputs at test time due to the distribution shift between training and test data~\cite{yang2021generalized, gawlikowski2021survey}.

\begin{figure}[t]
\centering
\includegraphics[
width=0.85\linewidth, trim={0cm 0cm 0cm 0cm},clip,
]{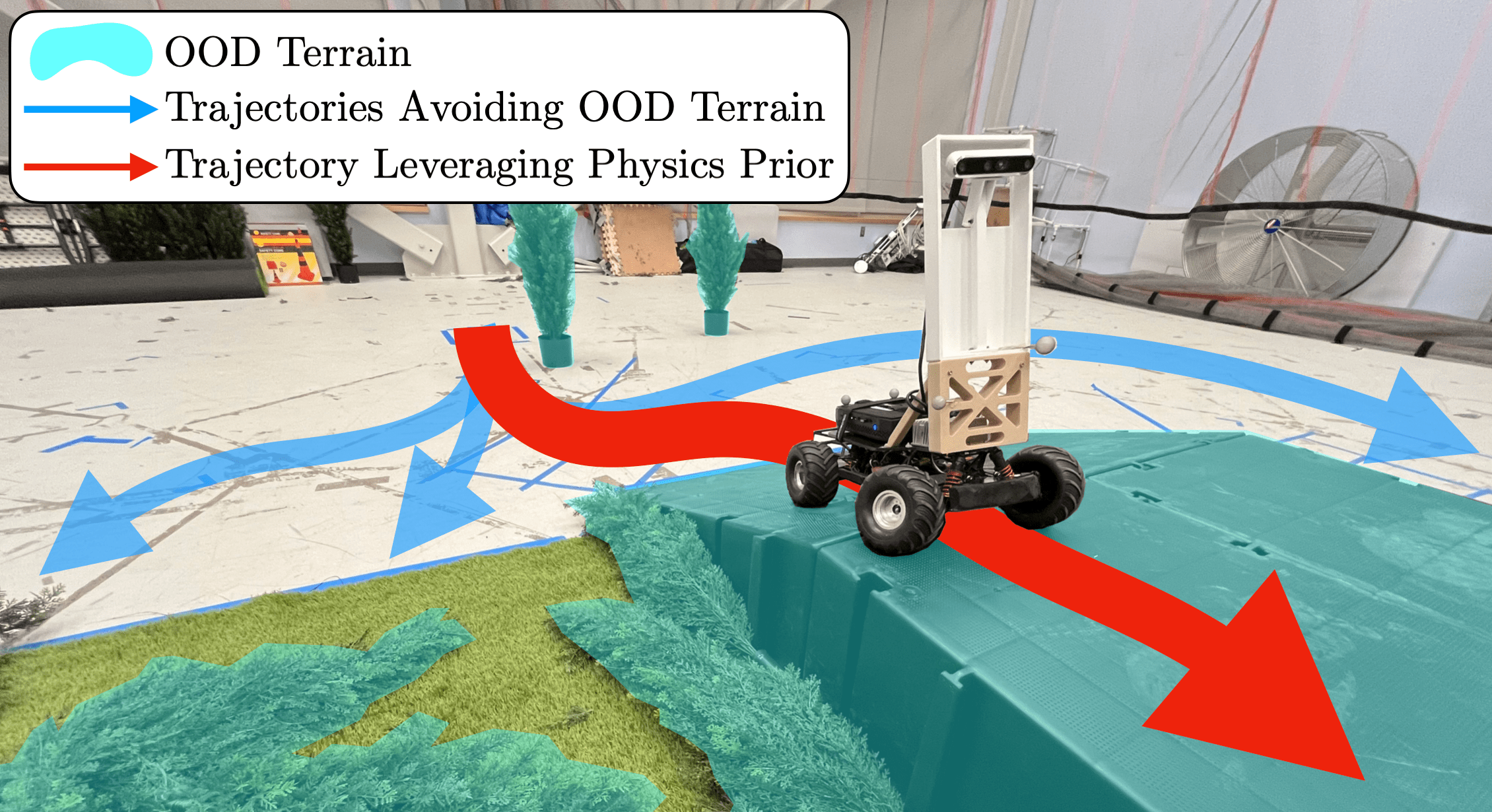}
\caption{
Real-world navigation scenario where the robot, trained on flat ground and turf, encounters unseen tall vegetation and ramps while navigating to a goal. Unlike prior works that avoid OOD terrain, this work successfully navigates across OOD terrain to reach the goal by integrating physics knowledge into the traversability model.
}
\label{fig:motivation}
\vspace*{-0.2in}
\end{figure}

Many recent works mitigate the risk of encountering OOD scenarios by quantifying the \textit{epistemic uncertainty}, which is the model uncertainty due to distribution shift~\cite{gawlikowski2021survey}. For example, OOD terrain can be detected by learning a density estimator for the training data to identify test input with low density~\cite{ancha2024icra, cai2023probabilistic}, or a terrain auto-encoder for detecting poorly reconstructed terrain~\cite{Frey-RSS-23, schmid2022self}. While avoiding OOD terrain has been shown to improve mission success rate in our prior work~\cite{cai2024evora}, doing so can be too conservative, such as the situation considered in this work (see Fig.~\ref{fig:motivation}). 
To this end, we propose \textbf{P}hysics-\textbf{I}nformed \textbf{E}vidential \textbf{TRA}versability (PIETRA), a self-supervised learning framework that seamlessly combines learning-based and physics-based traversability analysis methods such that the downstream planner relies on the learned model for in-distribution (ID) terrain and the physics-based model for OOD terrain. Improving upon our prior work~\cite{cai2024evora}, we exploit the mathematical structure of evidential learning to embed a physics-based prior that automatically gets invoked when epistemic uncertainty is high. Moreover, we propose an uncertainty-aware physics-informed loss function inspired by existing works to regularize the learned model to improve generalization~\cite{saviolo2022physics, maheshwari2023piaug}. 

\textbf{}In summary, the contributions of this work are threefold:
\begin{enumerate}
    \item An evidential traversability learning framework with an explicit physics-based prior that gets invoked when encountering OOD features at test time.
    \item An uncertainty-aware physics-inspired loss function that implicitly injects physics knowledge at training time to improve learning accuracy for both ID and OOD features.
    \item Extensive simulation and hardware experiments showing that our approach improves both the learning accuracy and the downstream navigation performance in test environments with significant distribution shift.
\end{enumerate}

\section{Related Work}\label{sec:related_work}
The field of traversability analysis studies how to infer suitability of terrain for navigation (see survey~\cite{papadakis2013terrain}).
Compared to hand-crafting planning costs based on terrain features, directly learning traversability models from data requires less manual labeling and results in a more accurate assessment of vehicle-terrain interaction. Based on navigation data, visual and/or geometric features of the visited terrain can be used to train a traversability predictor based on estimated traversability values~\cite{datar2024terrain, Frey-RSS-23, datar2023learning, cai2022risk, sikand2022visual}. Building upon this basic idea, visual and visual-inertial representation learning~\cite{seo2023learning, pokhrel2024cahsor}, self-training with pseudo-labels for unvisited terrain~\cite{cho2024learning}, temporal fusion of robot states and sensor measurements~\cite{gasparino2024wayfaster}, and data augmentation via vision foundation models~\cite{jung2024vstrong} can all improve learning accuracy. Alternatively, when a hand-crafted traversability model such as~\cite{fan2021step} is available, it can be used to provide supervisory signals for training NNs that are faster~\cite{frey2024roadrunner, triest2024unrealnet}. 
One concern is that, because learning-based methods rely on likely limited real-world data, the learned models may not generalize to environments unseen during training.
While our work also uses self-supervised learning to obtain a traversability model, we incorporate physics knowledge into the model to improve generalization to OOD environments.

OOD detection is well studied and closely related to uncertainty quantification (see surveys~\cite{yang2021generalized, gawlikowski2021survey}). At a high level, input features that are not well-represented in the training data can lead to high \textit{epistemic uncertainty}, which can be estimated with techniques such as Bayesian dropout~\cite{gal2016dropout}, model ensembles~\cite{osband2016deep}, and evidential methods~\cite{ulmer2023prior}. Epistemic uncertainty can also be estimated via a terrain auto-encoder for detecting high reconstruction error~\cite{schmid2022self}, a density estimator fit to the training data distribution~\cite{cai2023probabilistic}, or Gaussian Process regression~\cite{endo2023risk}.
Our prior work~\cite{cai2024evora} adopts the evidential method proposed by~\cite{natpn} to efficiently identify OOD terrain and shows that avoiding OOD terrain during planning can improve mission success rate. However, always avoiding OOD terrain can be too conservative. %
To address this limitation, this new work exploits the mathematical formulation of evidential learning to explicitly embed a physics prior that is activated when encountering OOD features. 
While informative priors have recently been combined with evidential learning in other problem settings, such as the use of a rule-based prior for trajectory prediction in autonomous driving~\cite{patrikar2024rulefuser}, our work focuses on off-road navigation and additionally introduces a physics-informed loss to further improve OOD generalization.

Incorporating physics and expert knowledge for navigating challenging terrain is crucial for both performance and safety, which can be achieved \textit{explicitly} or \textit{implicitly} (see survey~\cite{hao2022physics}). For example, explicit safety constraints based on terrain geometry and robot states can be imposed during planning~\cite{han2023model, sharma2023ramp, talia2023hound, pokhrel2024cahsor}. 
Physics laws can be explicitly incorporated into NNs via differentiable physics engines or neuro-symbolic methods~\cite{agishev2023monoforce, zhao2024physord}.
In addition, custom models can be directly used as priors in an evidential framework~\cite{patrikar2024rulefuser}. 
In contrast to previously mentioned explicit approaches, physics knowledge can also be infused into NN models implicitly by learning to reduce prediction errors with respect to both the training data and the physics model~\cite{saviolo2022physics, maheshwari2023piaug}.
This work described herein uses both explicit and implicit methods to infuse physics knowledge into the learned traversability model. As will be shown in Sec.~\ref{sec:sim_results}, compared to NNs trained with a physics-based loss that suffer from distribution shifts in far-OOD regime, our method gracefully falls back to the explicit physics prior. Furthermore, compared to an evidential network that only has a physics prior, our method uses a physics-based loss to further improve generalization.

\section{Problem Formulation}
We consider the problem of motion planning over uneven terrain for a ground vehicle. This section introduces the robot model with uncertain traversability parameters caused by rough terrain and sensing uncertainty in Sec.~\ref{sec:dynamical_models} and the risk-aware planning formulation in Sec.~\ref{sec:risk_aware_planning}. Compared to our prior work~\cite{cai2024evora} that only considers the linear and angular traction, this new work additionally accounts for the roll and pitch angles important for navigating uneven terrain.

\subsection{Dynamical Models with Traversability Parameters}\label{sec:dynamical_models}
Consider the discrete time system:
\begin{equation}\label{eq:dynamics}
    \mathbf{x}_{t+1} = F(\mathbf{x}_t, \mathbf{u}_t, \Param_{t}),
\end{equation}
where $\mathbf{x}_t\in\setX\subseteq\R^n$ is the state vector such as the position and heading of the ground robot, $\mathbf{u}_t\in\R^m$ is the control input, and $\Param_{t}\in\setParam\subseteq \R^r$ is the parameter vector that captures traversability of the terrain.
In this work, we focus on the bicycle model which is applicable for Ackermann-steering robots used in our simulation and hardware experiments:
\begin{equation}\label{eq:bicycle}
    \begin{bmatrix}
        p_{t+1}^{x}\\
        p_{t+1}^{y}\\
        \theta_{t+1}
    \end{bmatrix} = 
    \begin{bmatrix}
        p_{t}^{x}\\
        p_{t}^{y}\\
        \theta_{t}
    \end{bmatrix} + \Delta \cdot
    \begin{bmatrix}
        \psi_{1, t} \cdot v_t \cdot \cos{(\theta_t)}\\
        \psi_{1, t} \cdot v_t \cdot \sin{(\theta_t)}\\
        \psi_{2, t} \cdot v_t \cdot \tan (\delta_t)/L
    \end{bmatrix},
\end{equation}
where $\mathbf{x}_t=[p_{t}^{x}, p_{t}^{y}, \theta_{t}]\tr$ contains the X, Y positions and yaw angle; $\mathbf{u}_t=[v_t, \delta_t]\tr$ contains the commanded speed and steering angle; $0\leq \psi_{1,t}, \psi_{2,t} \leq1$ are the linear and angular traction; $\Delta>0$ is the time interval; and $L>0$ is the wheelbase. We additionally consider the absolute roll and pitch angles of the robot $\psi_{3,t}, \psi_{4, t} \geq 0$ that do not appear in~\eqref{eq:bicycle}. Therefore, the traversability parameter vector is $\Param_{t}=[\psi_{1,t}, \psi_{2, t}, \psi_{3,t}, \psi_{4, t}]\tr$.
Intuitively, traction captures the ``slip'' or the ratio between achieved and commanded velocities, which is important for fast navigation, and the roll and pitch values are important for rollover prevention. 

For rough terrain with vegetation, the traversability values are often unknown but can be empirically learned. Additionally, due to the noisy nature of the empirical data, we model traversability $\Param_{t}$ as random variables and mitigate the risk of encountering poor traversability during planning.

\subsection{CVaR-Based Risk-Aware Navigation}\label{sec:risk_aware_planning}
\begin{figure}[t]
	\centering
	\includegraphics[width=0.6\linewidth, trim={0cm 0cm 0cm 0cm},clip]{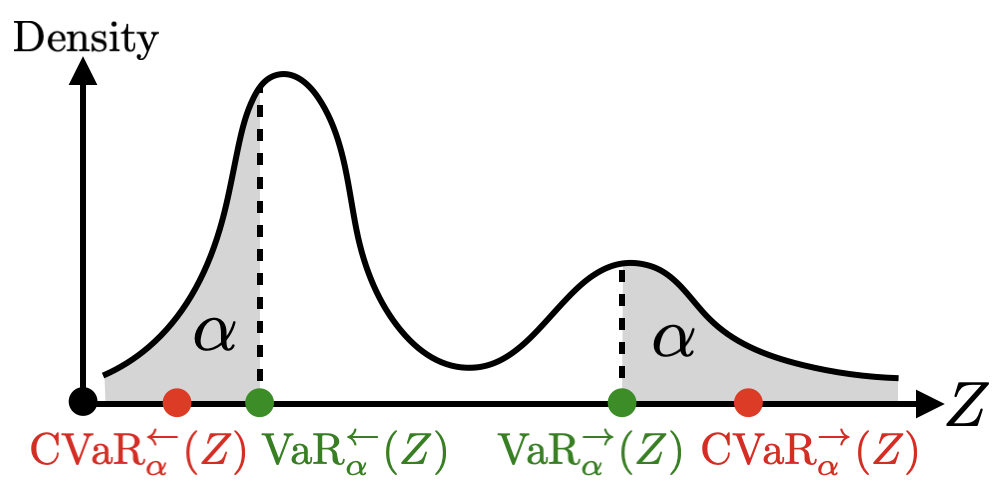}
 \caption{Conditional Value at Risk (CVaR) is the expected value of the worst-case $\alpha\in(0,1]$ portion of total probability and Value at Risk (VaR) is the worst-case $\alpha$-quantile for some random variable $Z$. We use both the left-tail and right-tail definitions proposed in~\cite{cai2024evora}.
 }\label{fig:cvar_definition}
\vspace*{-0.2in}
\end{figure}

Given the initial state $\mathbf{x}_0$ and maximum roll and pitch angles $\psi_{3}^{\max}, \psi_{4}^{\max}\geq 0$, we want to find a control sequence $\mathbf{u}_{0:T-1}$ that minimizes the time to reach the goal using the objective proposed in~\cite{cai2024evora}. We use Conditional Value at Risk (CVaR, as visualized in Fig.~\ref{fig:cvar_definition}) to quantify the risk of obtaining low traction and large roll and pitch angles. We achieve risk-aware planning by simulating the state trajectory using the left-tail $\text{CVaR}_\alpha^\leftarrow$ of traction and imposing the maximum attitude constraints over the right-tail $\text{CVaR}_\alpha^\rightarrow$ of roll and pitch angles:
\begin{align}
\min_{\mathbf{u}_{0:T-1}} \quad & C(\mathbf{x}_{0:T}) \label{eq:cvar_dyn}\\
\textrm{s.t.} \quad & \mathbf{x}_{t+1} = F(\mathbf{x}_t, \mathbf{u}_t, \bar{\Param}_{t})\\
&\bar{\psi}_{3, t} \leq \psi_{3}^{\max}, \quad \bar{\psi}_{4,t} \leq \psi_{4}^{\max} \\
&\bar{\Param}_{t}=
    \begin{bmatrix}
    \lcvar{\alpha}{\psi_{1,t}} \\
    \lcvar{\alpha}{\psi_{2,t}} \\
    \rcvar{\alpha}{\psi_{3,t}} \\
    \rcvar{\alpha}{\psi_{4,t}}
    \end{bmatrix},\ 
    \begin{bmatrix}
    \psi_{1,t} \\
    \psi_{2,t} \\
    \psi_{3,t} \\
    \psi_{4,t}
    \end{bmatrix} \sim p(\mathbf{o}_t)
    \\
&\mathbf{o}_t\text{ is the terrain feature at } \mathbf{x}_t\\
&\forall t\in\{0,\dots,T-1\} \label{eq:cvar_dyn_end},
\end{align}
where $\bar{\Param}_{t}$ contains the worst-case expected traversability values, $\alpha\in(0,1]$ is the risk tolerance, and $p(\mathbf{o}_t)$ is the traversability distribution after observing the terrain feature at state $\mathbf{x}_t$.
We use Model Predictive Path Integral control (MPPI~\cite{williams2017information}) to solve (\ref{eq:cvar_dyn}--\ref{eq:cvar_dyn_end}) because MPPI is gradient-free and parallelizable on GPU. Note that a similar formulation of (\ref{eq:cvar_dyn}--\ref{eq:cvar_dyn_end}) has been shown by~\cite{cai2024evora} to outperform methods that assume no slip and state-of-the-art methods such as~\cite{wang2021adaptive, cai2022risk}, but this new work  introduces additional constraints over roll and pitch.

\section{Physics-Informed Evidential Learning}\label{sec:pietra}
\begin{figure}[t]
\centering
\includegraphics[
width=\linewidth, trim={0cm 0cm 0cm 0cm},clip,
]{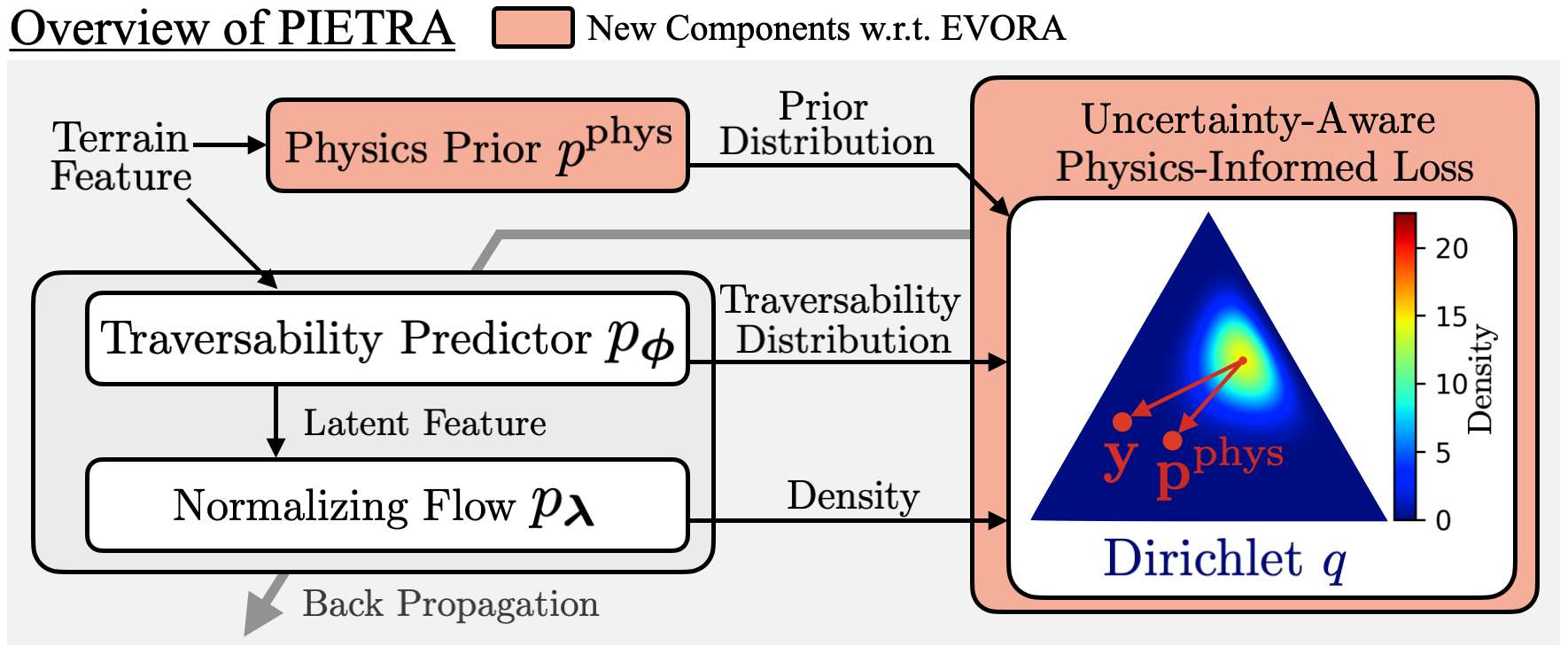}
\caption{
Overview of the proposed physics-informed evidential learning framework. 
In contrast to our prior work EVORA~\cite{cai2024evora}, this work explicitly embeds a physics prior in the Dirichlet posterior update and implicitly infuses physics knowledge via an uncertainty-aware, physics-informed loss during training.
}
\label{fig:architecture}
\vspace*{-0.2in}
\end{figure}

In this section, we present the proposed method PIETRA shown in Fig.~\ref{fig:architecture}. At a high level, the traversability predictor $p_{\bm{\phi}}$ parameterized by $\bm{\phi}$ outputs a categorical distribution over discretized traversability values to capture aleatoric uncertainty (inherent, irreducible data uncertainty) for each traversability parameter. Moreover, the normalizing flow network~\cite{Kobyzev2021flow} $p_{\bm{\lambda}}$ parameterized by $\bm{\lambda}$ estimates the densities of latent features as proxies for epistemic uncertainty (model uncertainty due to distribution shift). 
PIETRA improves upon our prior work EVORA~\cite{cai2024evora} by having an explicit physics prior that is invoked when encountering OOD inputs (Sec.~\ref{sec:explicit_physics_prior}) and an uncertainty-aware physics-informed loss that implicitly injects physics knowledge to further improve model accuracy in OOD terrain (Sec.~\ref{sec:implicit_physics_loss}).  Lastly, we design custom physics priors used in our experiments in Sec.~\ref{sec:phys_prior_design} and discuss the implementation details in Sec.~\ref{sec:impl_details}.

\subsection{Dirichlet Distribution with Physics Prior}\label{sec:explicit_physics_prior}
The Dirichlet distribution $q=\dir(\bm{\beta})$ with concentration parameters $\bm{\beta}=[\beta_1,\dots,\beta_B]\tr \in\R^{B}_{> 0}$  is a hierarchical distribution over categorical distributions $\cat(\mathbf{p})$, where $\mathbf{p}\in\R^{B}_{\geq0}$ is a normalized probability mass function (PMF) over $B>0$ discretized traversability values, i.e., $\sum_{b=1}^B p_b = 1$. The parameters $\mathbf{p}$ of the lower-level categorical distribution $\mathrm{Cat}(\mathbf{p})$ are sampled from the higher-level Dirichlet distribution, i.e., $\mathbf{p}\sim\dir(\bm{\beta})$. The mean (also called the \textit{expected PMF}) of the Dirichlet distribution is given by $\mathbb{E}_{\mathbf{p} \sim q} \left[ \mathbf{p} \right] = \bm{\beta}/\sum_{b=1}^{B} \beta_b$. 
The sum $n=\sum_{b=1}^{B} \beta_b$ reflects how concentrated the Dirichlet distribution is around its mean and corresponds to the ``total evidence'' of a data point observed during training.

Given an input feature $\mathbf{o}$ and a physics model $p^\text{phys}$ that maps its input to a traversability PMF with the evidence $n^\text{phys}>0$, the NN performs an input-dependent posterior update: 
\begin{align}
\bm{\beta}_{\bm{\phi}, \bm{\lambda}}^{\mathbf{o}} &= n^\text{phys} p^\text{phys}(\mathbf{o}) + n_{\bm{\lambda}}^{\mathbf{o}} p_{\bm{\phi}}(\mathbf{o})\label{eq:dirichlet_posterior}, \\
n_{\bm{\lambda}}^{\mathbf{o}} &=Np_{\bm{\lambda}}(\mathbf{z}_{\mathbf{o}})\label{eq:dir_new_evidence},
\end{align}
where the posterior Dirichlet distribution $q_{\bm{\phi}, \bm{\lambda}}^{\mathbf{o}} = \dir(\bm{\beta}_{\bm{\phi}, \bm{\lambda}}^{\mathbf{o}})$ depends on the physics prior and its evidence, the predicted traversability PMF $p_{\bm{\phi}}(\mathbf{o})$ and the predicted evidence $n_{\bm{\lambda}}^{\mathbf{o}}$ that is proportional to the density $p_{\bm{\lambda}}(\mathbf{z}_{\mathbf{o}})$ for the latent feature $\mathbf{z}_{\mathbf{o}}$ weighted by a fixed certainty budget $N>0$.
The posterior Dirichlet distribution leads to the expected traversability PMF:
\begin{equation}\label{eq:phys_prior_in_posterior_update}
    \mathbf{p}^{\mathbf{o}}_{\bm{\phi}, \bm{\lambda}} = \frac{n^\text{phys}p^\text{phys}(\mathbf{o})+ n_{\bm{\lambda}}^{\mathbf{o}} p_{\bm{\phi}}(\mathbf{o}) }{n^\text{phys}+  n_{\bm{\lambda}}^{\mathbf{o}} },
\end{equation}
which is used by the risk-aware planner (\ref{eq:cvar_dyn}--\ref{eq:cvar_dyn_end}).
The prior evidence is set to a small number (e.g., $n^\text{phys}=B$) such that the predicted evidence $n_{\bm{\lambda}}^{\mathbf{o}}$ is much larger than $n^\text{phys}$ for ID features. 
As a result, the learned traversability PMF $p_{\bm{\phi}}(\mathbf{o})$ is used for planning on ID terrain.
However, the new evidence $n^{\mathbf{o}}_{\bm{\lambda}}$ diminishes if the features are OOD, so the physics model $p^\text{phys}(\mathbf{o})$ is used for planning instead. The graceful transition between learned and physics-based traversability estimates eliminates the need to avoid OOD terrain, as is required in our prior work EVORA~\cite{cai2024evora} that uses an uninformative prior.

\subsection{Uncertainty-Aware Physics-Informed Loss }\label{sec:implicit_physics_loss}
In addition to explicitly embedding the physics prior in the posterior update~\eqref{eq:phys_prior_in_posterior_update}, physics knowledge can also be implicitly infused into the NN by adding a physics-based loss term during training. In contrast to existing methods~\cite{saviolo2022physics, maheshwari2023piaug} that are not uncertainty-aware, we adapt the physics-informed loss to the context of evidential learning.

We design the training loss based on the squared Earth Mover's distance ($\emdsq$~\cite{hou2016emd2}) which is a better measure of error compared to KL divergence that treats prediction errors independently across discrete bins. $\emdsq$ has a closed-form expression after dropping the constant multiplicative factor:
\begin{equation}\label{eq:emd2}
    \emdsq(\mathbf{p}, \mathbf{y}) \defeq \| \cumsum(\mathbf{p}) - \cumsum(\mathbf{y})  \|^2 ,
\end{equation}
where $\cumsum:\R^B\rightarrow \R^B$ is the cumulative sum operator,  and $\mathbf{p}$ and $\mathbf{y}$ are the predicted and target PMFs. 
To train an evidential NN, our prior work proposes the uncertainty-aware $\emdsq$ ($\uemdsq$) loss, which is the expected $\emdsq$ under the predicted Dirichlet $q$ (see the closed form in~\cite[Thm.~1]{cai2024evora}):
\begin{equation}
L^{\uemdsq}(q, \mathbf{y}) \defeq \E_{\mathbf{p}\sim q}~\big[ \emdsq (\mathbf{p}, \mathbf{y}) \big].\label{eq:uemd2_def}    
\end{equation}
This uncertainty-aware loss has been shown to improve both learning accuracy and navigation performance compared to using~\eqref{eq:emd2} or cross-entropy-based losses for training.

To implicitly infuse physics knowledge, we propose the uncertainty-aware physics-informed (UPI) loss,  defined as the expectation of the weighted sum of data loss and physics loss given the predicted Dirichlet distribution:
\begin{align}
L^{\text{UPI}}(q, \mathbf{y}) \nonumber &\defeq \E_{\mathbf{p}\sim q}~\big[ \emdsq (\mathbf{p}, \mathbf{y}) + \kappa \cdot  \emdsq (\mathbf{p}, \mathbf{p}^{\text{phys}})  \big] \\
&= \uemdsq (q, \mathbf{y}) + \kappa \cdot  \uemdsq (q, \mathbf{p}^{\text{phys}}),\label{eq:upi}
\end{align}
where $\mathbf{p}^{\text{phys}}=p^\text{phys}(\mathbf{o})$ is the PMF predicted by the physics model given some feature $\mathbf{o}$, and $\kappa\geq0$ is a hyperparameter. Intuitively, the physics-based term ensures that NN predictions stay close to the physics prior, thus improving generalization in near-OOD regimes. As our ablation study in Sec.~\ref{sec:sim_ablation} later shows, both the UPI loss~\eqref{eq:upi} and the physics prior~\eqref{eq:phys_prior_in_posterior_update} are important for achieving the best learning performance.

\subsection{Uncertainty-Aware Physics Prior}\label{sec:phys_prior_design}
We now present our custom physics priors for uneven terrain with two semantic types: the rigid \textit{dirt} terrain and the soft \textit{vegetation} terrain. Note that alternative prior designs could be used provided they are based on physics principles that hold in both ID and OOD scenarios.

\begin{figure}[t]
\centering
\includegraphics[
width=\linewidth, trim={0cm 0cm 0cm 0cm},clip,
]{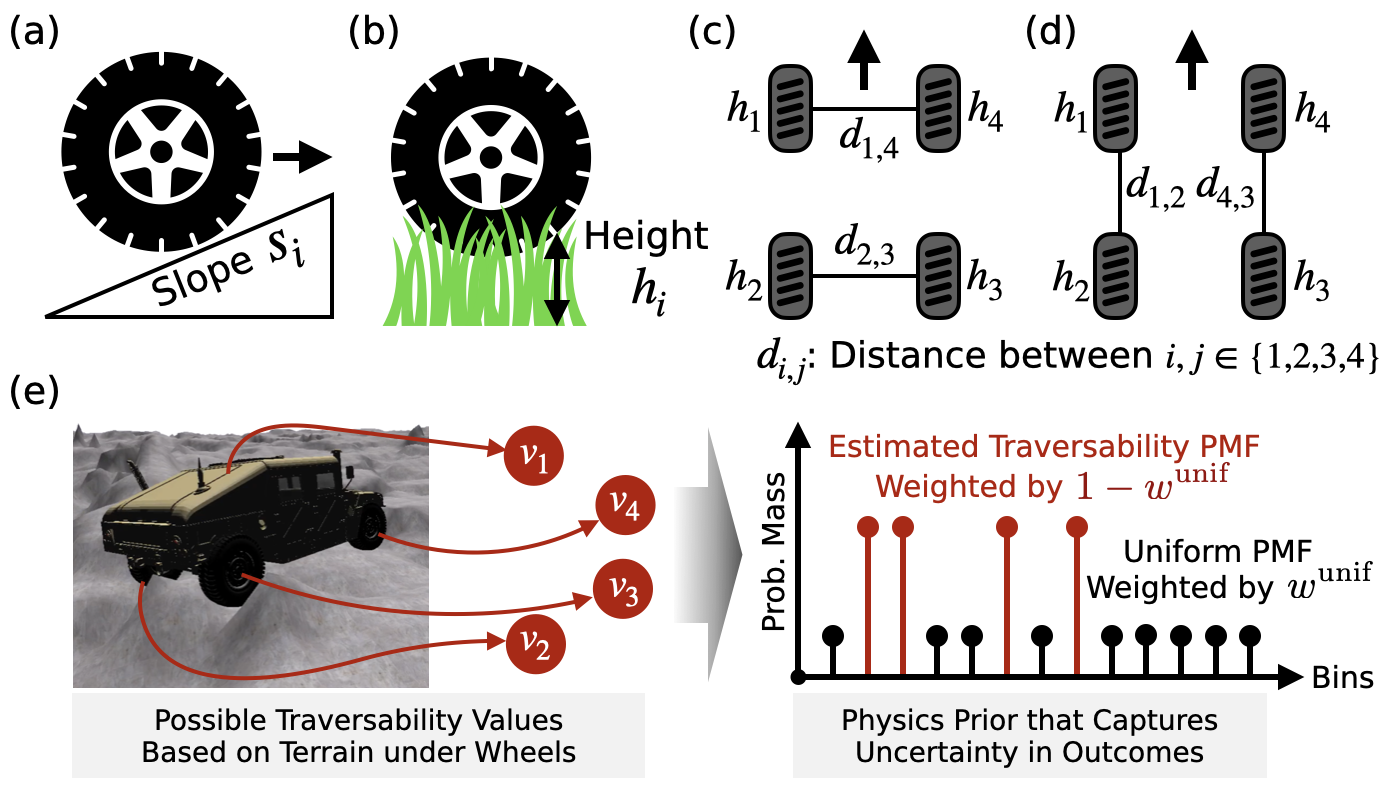}
\caption{
Terrain features used by the physics priors and how multiple potential traversability values are fused, as discussed in Sec.~\ref{sec:phys_prior_design}.
(a)~Terrain slope.
(b)~Terrain height.
(c)~Inter-wheel distances of the front wheel pair and back wheel pair. 
(d)~Inter-wheel distances of the left wheel pair and right wheel pair. 
(e)~Multiple candidates are fused into a single estimated traversability PMF and combined with the uniform distribution to account for uncertainty.
}
\label{fig:physics_prior}
\vspace*{-0.2in}
\end{figure}

The traction priors are inferred from the terrain properties under the wheels (also called the ``footprint'') assuming the robot has $0^\circ$ roll and pitch. 
Over dirt terrain, as the tire traction depends on friction that reduces with greater slopes, we design a simple slope-based model that predicts lower traction for more sloped terrain.
In our hardware experiment, taller vegetation deforms less and slows down the robot more, so we propose a traction model that predicts lower traction for taller vegetation.
To factor in the uncertainty due to outcomes from the four wheels, our physics-based traction models for the traversability parameter $\psi'\in\{\psi_1, \psi_2\}$ are
\begin{align}
\mathbf{p}^{\psi'}_{\text{dirt}} = 
     &\frac{1}{4} \sum_{i=1}^4 \mathbb{1}^{\psi'}_{\text{dirt}} \left( \text{clip}\left( \frac{s^{\max} - s_i}{s^{\max}}, 0, 1\right) \right) \label{eq:dirt_traction_prior}, \\
\mathbf{p}^{\psi'}_{\text{veg}} = 
     &\frac{1}{4} \sum_{i=1}^4 \mathbb{1}^{\psi'}_{\text{veg}} \left( \text{clip}\left( \frac{h^{\max} - h_i}{h^{\max}}, 0, 1\right) \right) \label{eq:veg_traction_prior},
\end{align}
where $s_i, h_i$ are the terrain slope and the height of vegetation under the wheel $i$, as shown in Fig.~\ref{fig:physics_prior}(a-b); the clip function restricts the traction values between 0 and 1; 
and the operator $\mathbb{1}^{\psi'}_{s}$ maps a scalar to a PMF where the estimated traversability value for $\psi'$ has probability 1 for the semantic type $s\in\mathcal{S}\defeq\{\text{dirt},\text{veg}\}$. 
Intuitively, the priors merge the estimated traction outcomes from the four wheels into a common PMF. Note that the terrain slope is estimated in the direction of the robot's heading, and we use the absolute slope to produce low traction values for both uphill and downhill terrain. 

The roll and pitch angles are estimated based on the terrain height under the wheels using trigonometry. The physics model for the traversability parameter $\psi'\in\{\psi_3, \psi_4\}$ and the semantic type $s\in\mathcal{S}$ is
\begin{equation}
\mathbf{p}^{\psi'}_s = 
    \frac{1}{2} \sum_{(i,j) \in \mathcal{W}^{\psi'} } \mathbb{1}_s^{\psi'} \left(  \arctan \left( \frac{h_i-h_j}{d_{i,j}} \right)  \right) \label{eq:attitude_prior},
\end{equation}
where $d_{i,j}$ is the distance between the wheels $i,j$. Roll and pitch priors use different wheel index pairs defined in $\mathcal{W}^{\psi_3}=\{(1,4), (2,3)\}$ and $\mathcal{W}^{\psi_4}=\{(1,2), (4,3)\}$ to account for uncertain terrain contact points as shown in Fig.~\ref{fig:physics_prior}(c-d).

To handle different semantic types within the footprint feature $\mathbf{o}$, the proposed physics prior for the traversability parameter $\psi'\in\{\psi_1, ..., \psi_4\}$ combines the prior predictions based on different semantic types via:
\begin{equation}\label{eq:phys_prior_with_uniform_baseline}
    p^{\text{phys}, \psi'}(\mathbf{o}) = w^\text{unif}\mathbf{p}^\text{unif} + (1-w^\text{unif})\sum_{s\in\mathcal{S}} r_s \mathbf{p}^{\psi'}_{s},
\end{equation}
where $r_s$ is the ratio of the semantic types within the footprint, $\mathbf{p}^{\psi'}_{s}$ is the physics prior based on (\ref{eq:dirt_traction_prior}--\ref{eq:attitude_prior}), $\mathbf{p}^\text{unif}$ is a uniform PMF, and $w^\text{unif}\in[0,1]$ determines how much uniform PMF is incorporated into the final physics prior to account for inaccurate prior predictions, as shown in Fig.~\ref{fig:physics_prior}(e). Note that~\eqref{eq:phys_prior_with_uniform_baseline} is used as the physics prior in~\eqref{eq:phys_prior_in_posterior_update}.

\subsection{Implementation Details}\label{sec:impl_details}
The traversability predictor $p_{\bm{\phi}}$ and the normalizing flow $p_{\bm{\lambda}}$ are trained jointly using the proposed UPI loss~\eqref{eq:upi} on an empirically collected dataset $\{(\mathbf{o}, \Param)_k\}_{k=1}^{K}$ where $K>0$ and every estimated traversability parameter in $\Param$ is converted to a target PMF $\mathbf{y}$ using one-hot encoding with $B=12$ discrete bins. Terrain features can be extracted from a semantic elevation map (e.g.,~\cite{erni2023mem}), while traversability values can be computed using the pose and velocity estimates obtained from onboard odometry, along with the commanded velocities.

For simplicity, we use a multi-layer perceptron (MLP) as the shared encoder to process the flattened and concatenated semantic and elevation patches of the terrain. The shared encoder is followed by separate MLP decoders with soft-max outputs for predictions of each traversability parameter. To reduce computation, we train a single flow network using the encoder's outputs. 
As recommended by~\cite{natpn}, we scale the constant certainty budget $N$ exponentially with the dimension of the encoder's outputs for numerical purposes. 
As the accuracy of physics priors differs across traversability parameters, we use a standalone MLP to map the shared encoder's output to scalars between 0 and 1 to downscale the predicted evidence in the posterior update~\eqref{eq:phys_prior_in_posterior_update} for each traversability parameter. 

In contrast to our prior work~\cite{cai2024evora}, this work uses yaw-aligned features because roll, pitch and traction are directional over uneven terrain. At deployment time, feature patches are obtained in a sliding-window fashion with a fixed number of yaw angles. The predicted traversability distributions are then converted to CVaR values for look-up during planning.

\section{Simulation Results}\label{sec:sim_results}

We use the Chrono Engine~\cite{chrono2016} to simulate navigation over rocky terrain and collect benchmark data. When compared against several baselines with or without physics knowledge, PIETRA achieves the best prediction accuracy (Sec.~\ref{sec:sim_learning_results}). Furthermore, we conduct an ablation study to validate the proposed improvements (Sec.~\ref{sec:sim_ablation}). When used for navigation, PIETRA leads to the best success rate and time to goal (Sec.~\ref{sec:sim_nav_results}). While only the dirt terrain is simulated, the real-world experiments have both dirt and vegetation (Sec.~\ref{sec:hw_results}).

\subsection{Simulation Setup}

\begin{figure}[t]
\centering
\includegraphics[
width=\linewidth, trim={0cm 0cm 0cm 0cm},clip,
]{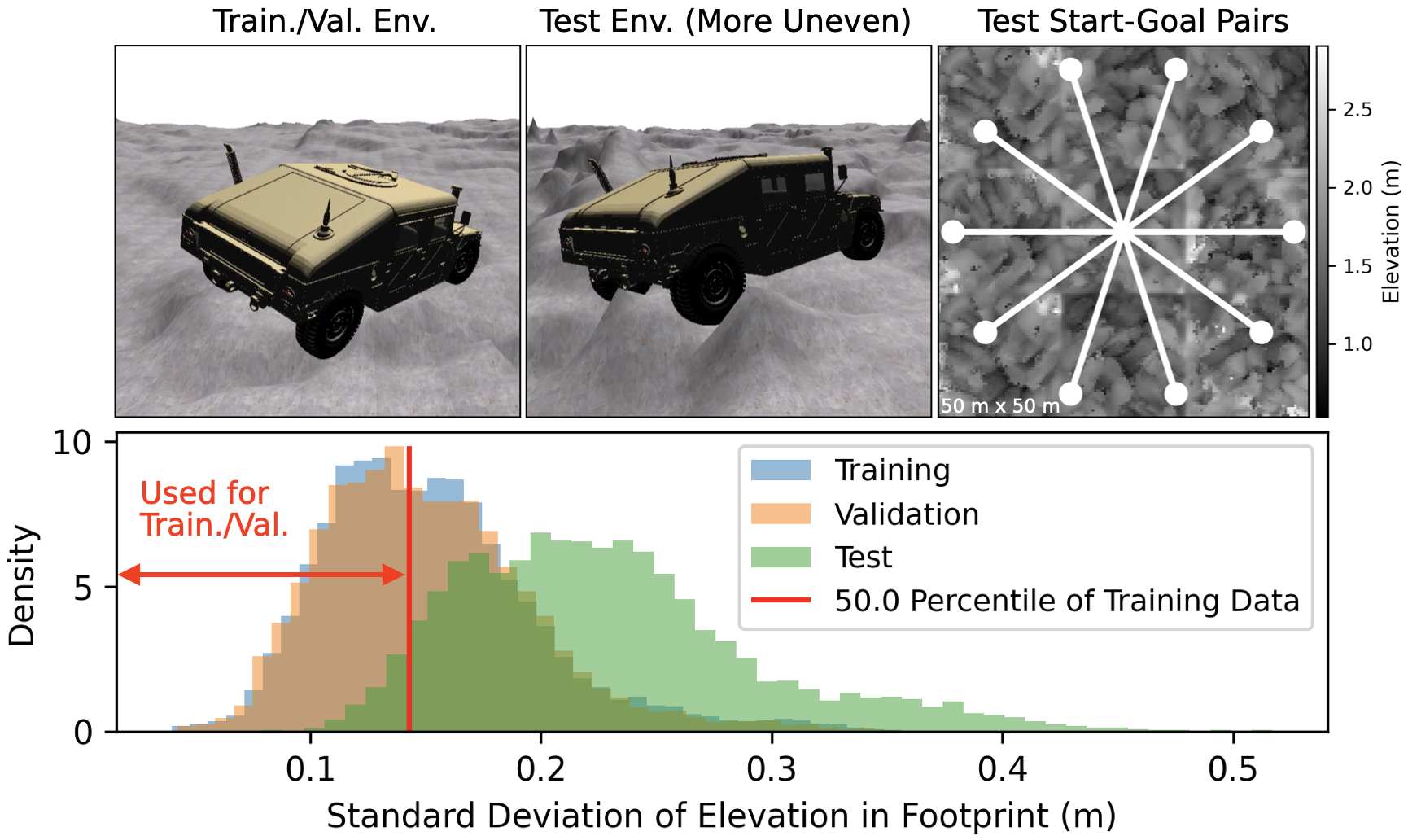}
\caption{Overview of the simulation setup with synthetic rocky terrain, where the terrain unevenness is measured by the standard deviation of elevation in the robot footprint. The test maps are more uneven than the training and validation ones to induce distribution shift. During training and validation, only the data with terrain unevenness below the 50th percentile of the training dataset is used. Note that test maps are also used to evaluate navigation performance.}
\label{fig:data_distribution}
 \vspace*{-0.2in}
\end{figure}

An overview of the simulation environment is shown in Fig.~\ref{fig:data_distribution}. We generate 30 distinct synthetic elevation maps based on the real-world rock testbed proposed in~\cite{datar2024toward}, which are split evenly for training, validation and testing. Every map has sides of $50$~m and we use an Ackermann-steering robot of dimension $3.4$~m by $1.8$~m to collect 20 navigation trials in each map. For simplicity, the robot executes sinusoidal steering and $2$~m/s speed commands, and the episode ends if the robot rolls over, gets stuck, or exits the map. Note that terrain features and traversability values are obtained directly from the simulator.

To emphasize the testing of OOD generalization, the test elevation maps are scaled more than the training and validation maps, inducing significant distribution shift as visualized in Fig.~\ref{fig:data_distribution}, where we use the standard deviations of the elevation features to measure terrain unevenness. 
To clearly mark the ID/OOD boundary, we consider a terrain feature as ID (or OOD) if its unevenness falls below (or above) the 50th percentile of the training dataset. We only use the ID data for training and validation, but the accuracy of trained NNs is evaluated on the entire test dataset. Note that the test maps are also used for evaluating navigation performance.

\subsection{Learning Benchmark}\label{sec:sim_learning_results}

The proposed method PIETRA is compared to several state-of-the-art methods, including our prior work EVORA~\cite{cai2024evora}, an encoder-decoder NN trained with the $\emdsq$ loss (Vanilla), and an encoder-decoder NN trained with the physics-informed loss (PI) adapted from~\cite{saviolo2022physics} to use the $\emdsq$ loss. 
Based on the training data and vehicle characteristics, we tune the max slopes~\eqref{eq:dirt_traction_prior} for linear and angular traction to correspond to $30^\circ$ and $15^\circ$ with $w^\text{unif}=0.2$ for mixing in the uniform PMF~\eqref{eq:phys_prior_with_uniform_baseline}.
Hyperparameter sweeps are conducted over learning rates in $\{1\mathrm{e}{-3}, 1\mathrm{e}{-4},1\mathrm{e}{-5}\}$ for the Adam optimizer, and weights for the physics loss term in $\{0.1, 0.5, 1.0\}$. To encourage smoothness, we penalize Dirichlet entropy~\cite{natpn} with weights in $\{1\mathrm{e}{-3}, 1\mathrm{e}{-4},1\mathrm{e}{-5}\}$ for evidential NNs. We identify the best parameter set for each method based on the average validation $\emdsq$ errors over $5$ training seeds and report the mean and standard deviations of the prediction errors over the seeds.

\begin{figure}[t]
\centering
\includegraphics[
width=\linewidth, trim={0cm 0cm 0cm 0cm},clip,
]{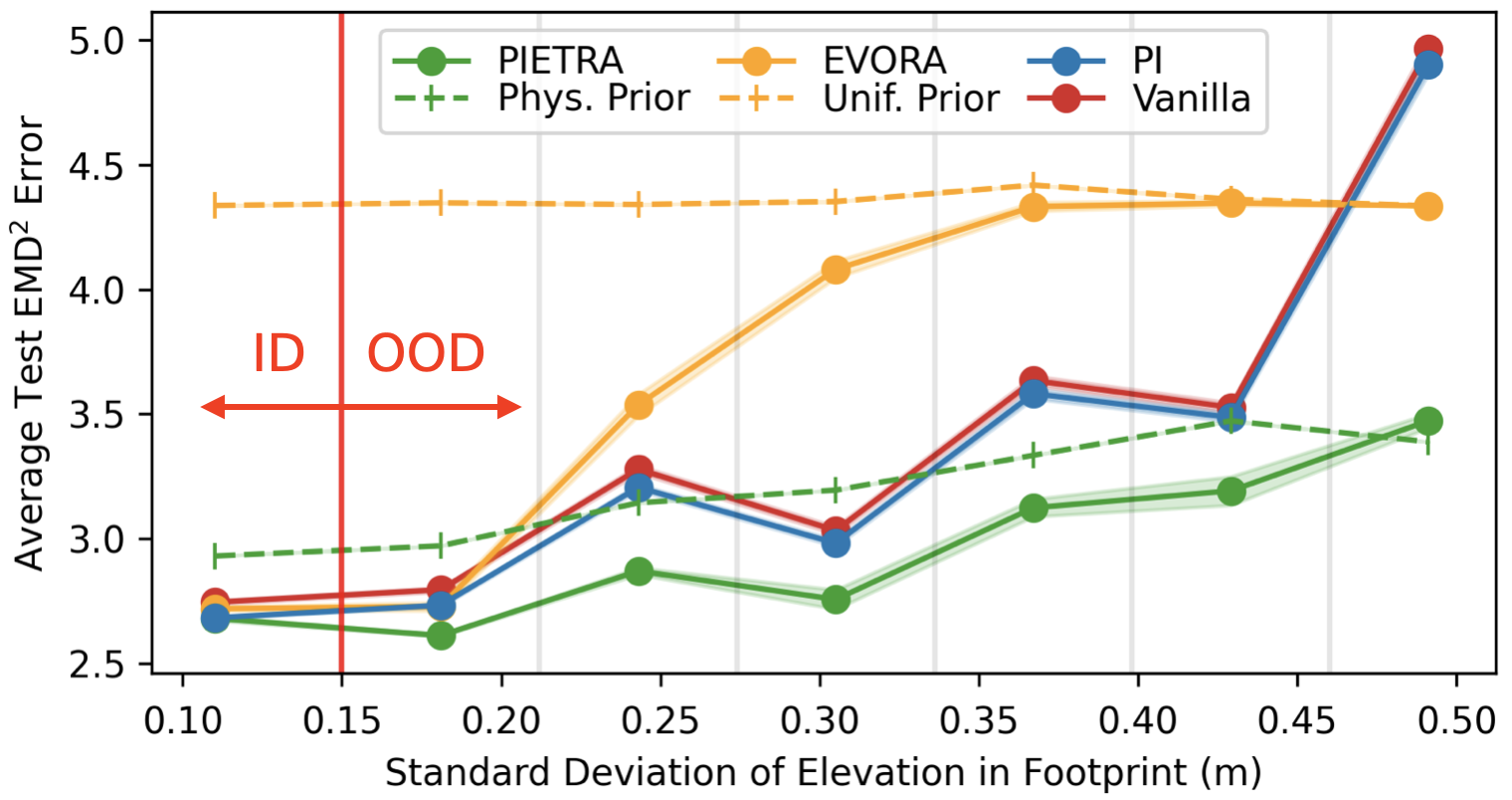}
\caption{Prediction errors binned by the standard deviation of elevation in robot footprint. While all trained models perform similarly on ID data, PIETRA generalizes the best to OOD data. Unlike PIETRA that falls back to the physics prior, EVORA falls back to the uniform prior which has higher error. Without explicit priors, PI and Vanilla perform poorly in far-OOD regime.}
\label{fig:nn_benchmark}
\vspace*{-0.2in}
\end{figure}

\begin{table}[b]
\centering
\setlength{\tabcolsep}{0.15cm}
\caption{Learning benchmark results.}
\label{tab:learning_benchmark}
\footnotesize{
\begin{tabular}{llll}
\toprule
                         & \multicolumn{3}{c}{Prediction Error ($\emdsq$ $\downarrow$)}                                          \\ \cmidrule(l){2-4}
\multirow{-2}{*}{Method} & Overall & ID                               & OOD                              \\ \midrule
\rowcolor[gray]{.9}
PIETRA (Proposed) & $\mathbf{2.77  \pm 0.01}$ & $\mathbf{2.68 \pm 0.01}$ & $\mathbf{2.78 \pm 0.01}$  \\
Physics-Informed (PI)  & $2.99 \pm 0.01$  & \cellcolor[gray]{.9}$\mathbf{2.68 \pm 0.01}$ & $3.02 \pm 0.01$ \\
Vanilla  & $3.05 \pm 0.01 $ & $2.74 \pm 0.01$  & $3.08 \pm 0.01$ \\ 
EVORA  & $3.30 \pm 0.02$  & $2.72 \pm 0.01$ & $3.35 \pm 0.02$ \\ \midrule
\rowcolor[gray]{.9}
Physics Prior   & $\mathbf{ 3.09}$  & $\mathbf{2.93}$ & $ \mathbf{3.10}$ \\ 
Uniform Prior  & $4.35 $  & $4.34$ & $4.35$ \\ \bottomrule
\end{tabular}
}
\end{table}
The learning benchmark results in Table~\ref{tab:learning_benchmark} show that PIETRA achieves the best overall, ID and OOD performance. To gain more intuition, the test errors are binned based on terrain unevenness in Fig.~\ref{fig:nn_benchmark}. Notably, while PI outperforms Vanilla and performs similarly to PIETRA on ID features, the improvement thanks to the physics-informed loss degrades as features become more OOD. On the other hand, PIETRA and EVORA fall back to their corresponding priors (physics prior and uniform prior) due to decreasing predicted evidence~\eqref{eq:phys_prior_in_posterior_update}.

\subsection{Ablation Study}\label{sec:sim_ablation}
An ablation study for the proposed improvements with respect to EVORA is summarized in Table~\ref{tab:ablation}. For completeness, we include a variant where EVORA uses the physics model for OOD features with latent densities lower than the densities observed during training. The takeaway is that both the physics prior and physics-informed loss are important for achieving the best overall accuracy. While the UPI loss~\eqref{eq:upi} alone leads to the best ID accuracy, it offers limited improvement on OOD accuracy. 
Interestingly, EVORA equipped with a physics prior achieves better OOD accuracy than OOD-based switching, verifying the benefit of the explicitly embedded physics prior.

\begin{table}[h]
\centering
\setlength{\tabcolsep}{0.15cm}
\caption{Ablation study on infusing physics knowledge into EVORA~\cite{cai2024evora}.}
\label{tab:ablation}
\footnotesize{
\begin{tabular}{llll}
\toprule
                         & \multicolumn{3}{c}{Prediction Error ($\emdsq$ $\downarrow$)}                                          \\ \cmidrule(l){2-4}
\multirow{-2}{*}{Changes w.r.t. EVORA}& Overall & ID                               & OOD                              \\ \midrule

\cellcolor[gray]{.9}PP \& UPI (Proposed) & \cellcolor[gray]{.9}$\mathbf{2.77  \pm 0.01}$ & $2.68 \pm 0.01$ & \cellcolor[gray]{.9}$\mathbf{2.78 \pm 0.01}$  \\
UPI  & $3.27 \pm 0.02$  & \cellcolor[gray]{.9}$\mathbf{2.67 \pm 0.01}$ & $3.32 \pm 0.02$ \\
PP  & $2.86 \pm 0.01$  & $2.73 \pm 0.01$ & $2.87 \pm 0.01$ \\
Use phys. model if OOD   & $2.93 \pm 0.01 $ & $2.72 \pm 0.01$  & $2.95 \pm 0.01$ \\ 
EVORA & $3.30 \pm 0.02$  & $2.72 \pm 0.01$ & $3.35 \pm 0.02$ \\ \bottomrule
\multicolumn{4}{l}{PP: Physics Prior\quad UPI: Uncertainty-Aware Physics-Informed Loss~\eqref{eq:upi}}
\end{tabular}
}
\vspace*{-0.2in}
\end{table}

\subsection{Navigation Benchmark}\label{sec:sim_nav_results}
We deploy the NNs from Sec.~\ref{sec:sim_learning_results} trained with the best hyperparameters in the test environments. As discussed in~\cite{cai2024evora}, EVORA detects OOD terrain that is avoided during planning via auxiliary costs. As an ablation of the physics prior, we also consider PIETRA with OOD avoidance. We consider 10 test maps and 10 start-goal pairs that are 40~m apart as visualized in Fig.~\ref{fig:data_distribution}. The robot aims to reach the goal without exceeding $30^\circ$ of roll and pitch angles or getting immobilized. The robot has a maximum steering angle of $30^\circ$ and a maximum speed of $1$~m/s to ensure stable crawling. Note that MPPI runs at 10~Hz on
GPU with a 5~s planning horizon and 1024 rollouts. We consider 3 risk tolerances $\alpha$ in $\{0.4, 0.6, 0.8\}$ and report the success rate and the time to goal of successful trials.

\begin{figure}[t]
\centering
\includegraphics[
width=\linewidth, trim={0cm 0cm 0cm 0cm},clip,
]{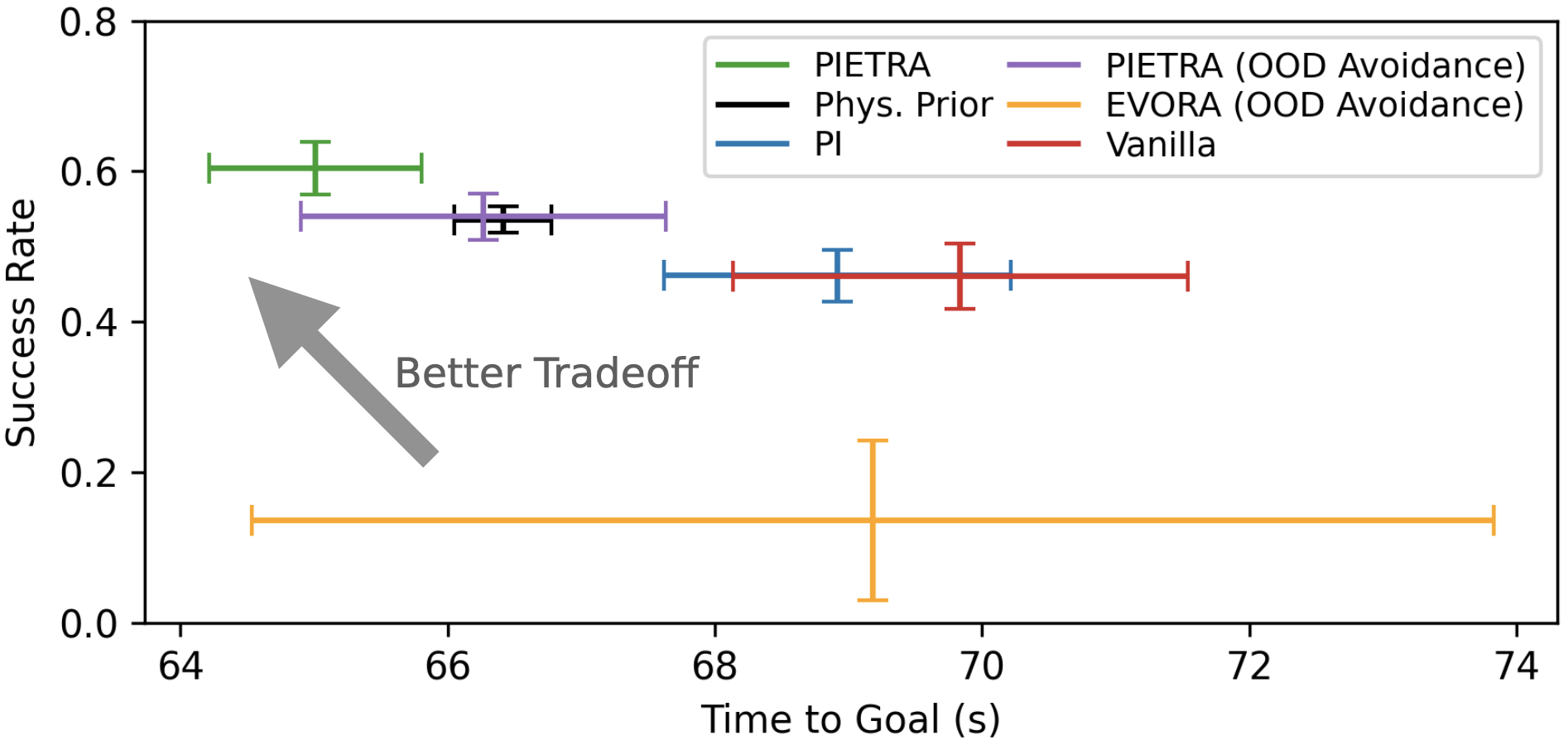}
\caption{
Success rate and time to goal from the simulated navigation benchmark. The figure shows the mean and standard deviation values for each method. The proposed PIETRA performs the best because it seamlessly transitions between the learned and physics models on OOD terrain. When PIETRA avoids OOD terrain, it becomes more conservative but still outperforms EVORA thanks to the physics-informed training loss. While PI outperforms Vanilla, it performs worse than the physics prior due to unreliable traversability predictions.
}
\label{fig:sim_nav_benchmark}
\vspace*{-0.2in}
\end{figure}

The navigation results in Fig.~\ref{fig:sim_nav_benchmark} show that PIETRA achieves the best success rate and time to goal. In comparison, PIETRA with OOD avoidance is more conservative but still outperforms EVORA thanks to the proposed UPI loss~\eqref{eq:upi}.
While PI outperforms Vanilla, it performs worse than the physics prior due to poor OOD generalization.

\section{Hardware Experiments}\label{sec:hw_results}
While the simulation results in Sec.~\ref{sec:sim_results} have shown improved learning and navigation performance achieved by PIETRA with respect to the state-of-the-art, this section presents further evidence of PIETRA's practical utility via real-world experiments on a computationally constrained platform in the presence of multiple semantic types in the environment.

\subsection{Data Collection, Training and Deployment}
The indoor 9.6~m by 8~m arena contains turf and fake bushes to mimic outdoor vegetation. The \textit{dirt} semantic type includes the concrete floor and a skateboard ramp.
A 0.33~m by 0.25~m RC car carries a RealSense D455 depth camera, and a computer with an Intel Core i7 CPU and Nvidia RTX 2060 GPU. The robot runs onboard traversability prediction, motion planning, and elevation mapping with 0.1~m resolution, but Vicon is used for estimating the pose and velocity of the robot. 
For simplicity, terrain elevation and semantics are estimated separately based on depth and color measurements. Each map cell contains a single semantic type initially set to dirt and is later updated to vegetation if the corresponding pixels in the image space are green.

\begin{figure*}[t]
\centering
\includegraphics[
width=\linewidth, trim={0cm 0cm 0cm 0cm},clip,
]{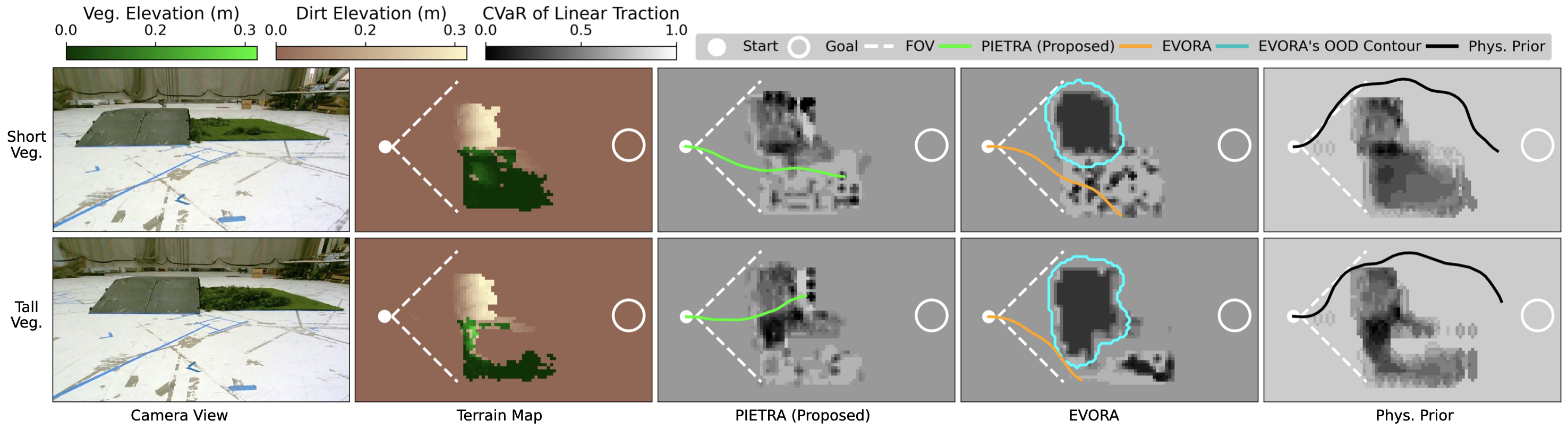}
\caption{
Snapshots of the start of the two missions with short and tall vegetation. From the onboard camera view, the robot builds semantic elevation map online, which is processed to generate CVaR of the traversability values. Due to limited space, we only show the CVaR of the predicted linear traction aligned with the robot heading for each method and overlay the initial best trajectories. For EVORA, we additionally show the OOD contour that the robot avoids.}
\label{fig:hw_exp_intuition}
\vspace*{-0.2in}
\end{figure*}

Traversability models are trained on 10~min of manual driving data over \textit{flat} concrete floor and turf, making the tall bushes and skateboard ramp OOD at test time. For the vegetation and dirt traction priors, we empirically tune the maximum dirt slope $s^{\max}$~\eqref{eq:dirt_traction_prior} to correspond to $30^\circ$ and the maximum vegetation height $h^{\max}$~\eqref{eq:veg_traction_prior} to be $0.2$~m which is slightly greater than the $0.15$~m wheel diameter. 
To prevent damaging the hardware, we do not consider PI and Vanilla because their predictions for OOD terrain are unreliable. Therefore, we only consider PIETRA, the physics prior, and EVORA that avoids OOD terrain. All models are trained with the learning rate of $1\mathrm{e}{-4}$, entropy weight of $1\mathrm{e}{-4}$ and physics loss weight of $0.1$ when applicable. We set a fixed risk tolerance of $\alpha=0.5$ for the risk-aware planner. Note that MPPI runs at 10~Hz on GPU with a 5~s planning horizon and 1024 rollouts, and the generation of CVaR maps runs at 5~Hz with the map dimension of $81\times 96$ with 13 yaw discretizations.

\subsection{Navigation Results}

\begin{figure}[t]
\centering
\includegraphics[
width=\linewidth, trim={0cm 0cm 0cm 0cm},clip,
]{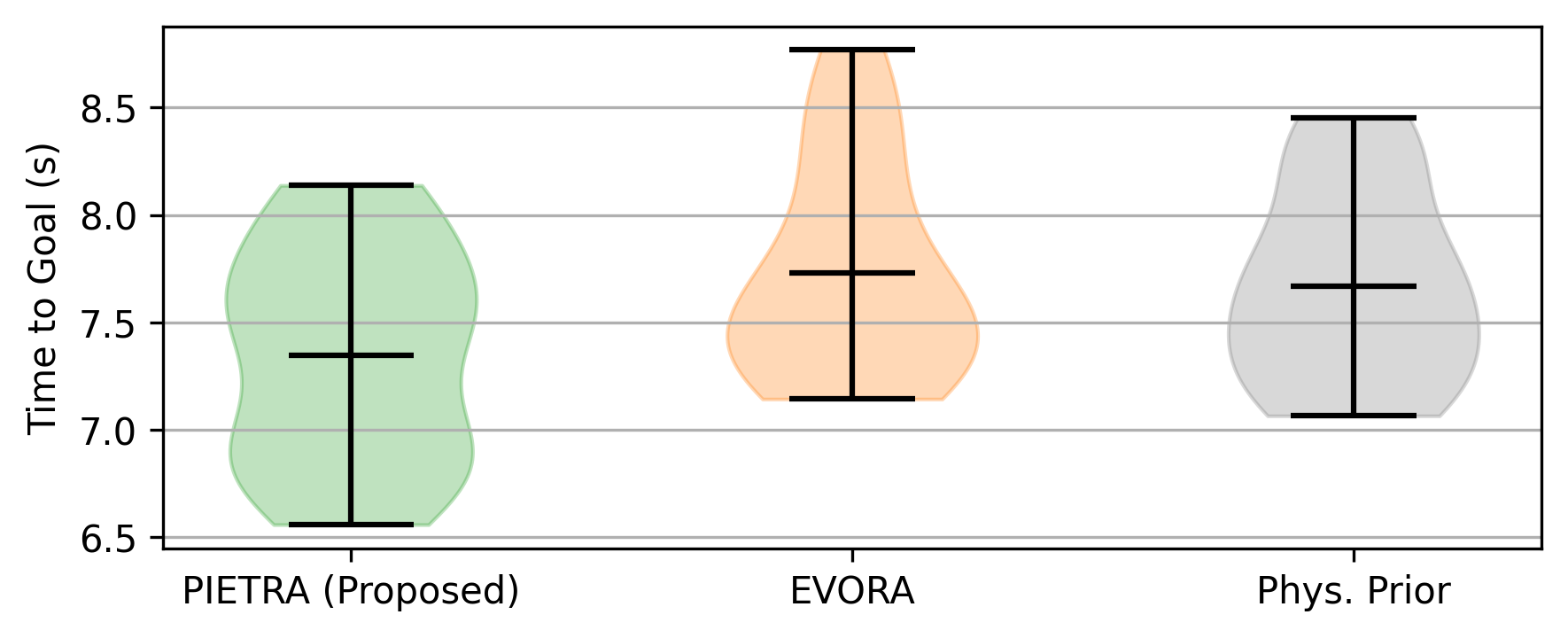}
\centering
\setlength{\tabcolsep}{0.15cm}
\footnotesize{
\begin{tabular}{llll}
\toprule
  &      & \multicolumn{2}{c}{Failure Rate by Type $\downarrow$}        \\ \cmidrule(l){3-4} 
\multirow{-2}{*}{Method} & \multirow{-2}{*}{Success Rate $\uparrow$} & Out of Bound & Roll Violation                   \\ \midrule
\rowcolor[gray]{.9} PIETRA (Proposed)  &        $\mathbf{19/20}$        &    $\mathbf{0/20}$     &      \cellcolor[HTML]{FFFFFF}$1/20$                        \\
EVORA        &              $17/20$          &   $3/20$      &   \cellcolor[gray]{.9}$\mathbf{0/20}$      \\
Physics Prior    &               $16/20$           &   $2/20$     &    $2/20$                         \\ \bottomrule
\end{tabular}
}
\caption{Hardware experiment results showing that PIETRA achieves the best time to goal and success rate. To avoid OOD terrain, EVORA takes wide turns that increase the time to goal and the odds of going out of bound. While the physics prior achieves better time to goal than EVORA, it suffers from more failures due to inaccurate traction estimates on the flat terrain.}
\label{fig:hw_results}
\vspace*{-0.2in}
\end{figure}

\begin{figure}[t]
\centering
\includegraphics[
width=\linewidth, trim={0cm 0cm 0cm 0cm},clip,
]{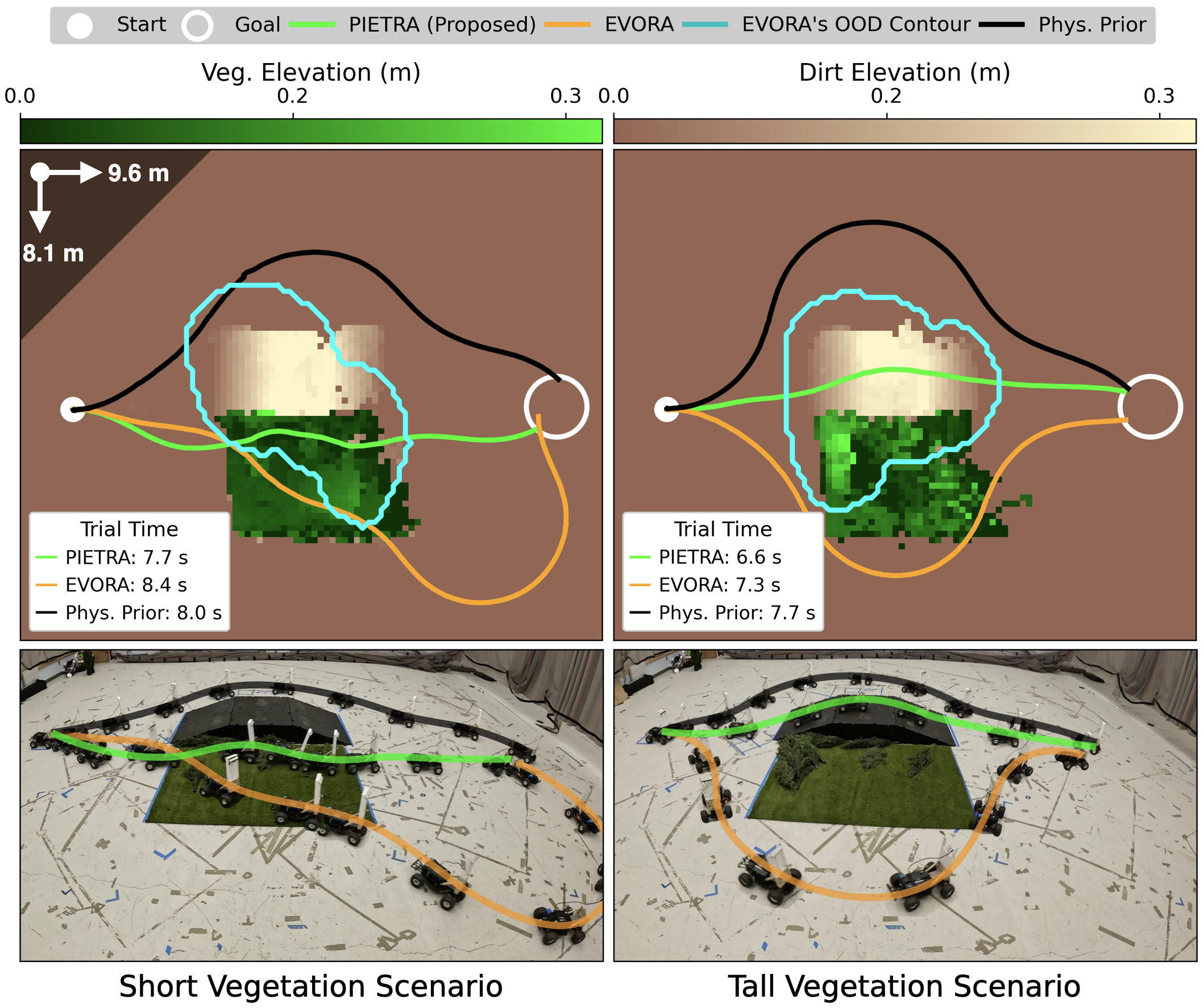}
\caption{Full trajectories, terrain maps and trial time of representative trials. PIETRA chooses between vegetation and the ramp based on vegetation height. While the physics prior and PIETRA share similar traversability assessment for tall vegetation and the ramp, the physics prior overestimates the terrain traction of the concrete floor, making the robot frequently go around the ramp and vegetation. EVORA detects and avoids the OOD terrain unseen during training, causing undesirable wide turns due to limited steering authority. }
\label{fig:hw_qualitative}
\vspace*{-0.2in}
\end{figure}

The robot is tasked to navigate to a fixed goal without leaving the bounding box of the arena or violating the roll and pitch constraints of $30^\circ$ and $45^\circ$ respectively, while deciding whether to drive on vegetation, climb over the ramp, or stay on the concrete floor. We vary the elevation and density of bushes and repeat each method 20 times. Some snapshots of the start of the missions are visualized in Fig.~\ref{fig:hw_exp_intuition} for scenarios with short and tall vegetation. The success rates and the time to goal of the successful trials are reported in Fig.~\ref{fig:hw_results}, showing that PIETRA achieves the best performance. PIETRA violates the roll constraint once because the robot drives off the down-ramp prematurely. 
EVORA has 3 out-of-bound failures because the robot frequently takes wide turns to avoid OOD terrain, which makes the robot more likely to miss the goal and leave the bounding box. The physics prior leads to 2 out-of-bound failures because it misses the goal region slightly due to over-optimistic traction estimates. The physics prior also leads to 2 roll violations due to driving off the down-ramp prematurely.

Representative trials in Fig.~\ref{fig:hw_qualitative} show that PIETRA goes over vegetation when the bushes are short, but it goes over the ramp when the bushes are tall. In comparison, the large OOD regions force EVORA to take wide turns. Interestingly, the robot using the physics prior favors the concrete floor. To explain this phenomenon, Fig.~\ref{fig:hw_exp_intuition} shows the predicted CVaR of linear traction for each method, suggesting that the physics prior is over-confident about the traction of the concrete floor. This further shows that PIETRA combines the best of EVORA and physics prior in that it relies on the learned model in ID terrain, but falls back to the physics model in OOD terrain. 
\section{Conclusion and Future Work}

We presented PIETRA, a self-supervised traversability learning method that incorporates physics knowledge explicitly via the custom physics prior and implicitly via the physics-informed training loss. Extensive simulations and hardware experiments show that PIETRA improves learning accuracy and navigation performance under significant distribution shifts. 

One of the limitations of PIETRA is its large memory footprint due to discretization, so improvements are needed for higher-dimensional systems. Moreover, PIETRA relies on accurate terrain features and state estimation, but the risk due to uncertain sensing and estimation may need to be addressed for some real-world environments. Lastly, PIETRA can be further improved by leveraging data augmentation~\cite{maheshwari2023piaug}, differentiable physics simulators~\cite{agishev2023monoforce} and neuro-symbolic networks~\cite{zhao2024physord}.

\bibliographystyle{IEEEtran}
\bibliography{bibs}

\end{document}